\documentclass{article}

\usepackage[preprint]{neurips_2019}

\usepackage[utf8]{inputenc} \usepackage[T1]{fontenc}    \usepackage{hyperref}       \usepackage{url}            \usepackage{booktabs}       \usepackage{amsfonts}       \usepackage{nicefrac}       \usepackage{microtype}      
\usepackage{amsmath}
\usepackage{enumitem}
\usepackage{color}

\usepackage[noend,linesnumbered,ruled]{algorithm2e}
\usepackage{graphicx}
\usepackage[font=small]{caption}
\usepackage[font=small]{subcaption}

\title{A Model-based Approach for Sample-efficient Multi-task Reinforcement Learning
}

\author{Nicholas C. Landolfi$^1$ \quad Garrett Thomas$^1$ \quad Tengyu Ma$^{1,2}$\\
\texttt{\{lando, gwthomas, tengyuma\}@stanford.edu} \\
\\
$^1$ Department of Computer Science, Stanford University\\
$^2$ Department of Statistics, Stanford University
}

\def\shownotes{1}  \ifnum\shownotes=1
\newcommand{\authnote}[2]{$\ll$\textsf{\footnotesize #1 notes: #2}$\gg$}
\else
\newcommand{\authnote}[2]{}
\fi

\DeclareMathOperator*{\maximize}{maximize}

\begin{document}

\maketitle

\newcommand\R{{\mathbb{R}}}
\newcommand\N{{\mathbb{N}}}
\newcommand\BE{{\mathbf E}}
\newcommand\CS{{\mathcal S}}
\newcommand\CA{{\mathcal A}}
\newcommand\CD{{\mathcal D}}
\newcommand\CL{{\mathcal L}}
\newcommand{\norm}[1]{\left\lVert#1\right\rVert}

\newcommand{\nicknote}[1]{{\textcolor{red}{\textbf{nick: #1}}}}
\newcommand{\dm}{M}
\newcommand{\ours}{METaL}

\begin{abstract}
The aim of multi-task reinforcement learning is two-fold: (1) efficiently learn by training against multiple tasks and (2) quickly adapt, using limited samples, to a variety of new tasks.
In this work, the tasks correspond to reward functions for environments with the same (or similar) dynamical models.
We propose to learn a dynamical model during the training process and use this model to perform sample-efficient adaptation to new tasks at test time.
We use significantly fewer samples by performing policy optimization only in a ``virtual" environment whose transitions are given by our learned dynamical model.
Our algorithm sequentially trains against several tasks.
Upon encountering a new task, we first \emph{warm-up} a policy on our learned dynamical model, which requires no new samples from the environment.
We then \emph{adapt the dynamical model} with samples from this policy in the real environment.
We evaluate our approach on several continuous control benchmarks and demonstrate its efficacy over MAML, a state-of-the-art meta-learning algorithm, on these tasks.
\end{abstract}

\section{Introduction}

Reinforcement learning has achieved significant success in domains such as game-playing \citep{atari}, recommender systems \citep{deeprecommenderssystems}, and robotic control \citep{visuomotor}.
When developing systems that must solve several tasks rather than a single task, it is natural to try to leverage shared structure across tasks to make the learning process more efficient \citep{caruana1997multitask}.
Moreover, exploiting shared structure enables faster adaptation to unseen tasks in the future.

The Model Agnostic Meta Learning (MAML) algorithm  has proven to be a successful algorithm for multi-task learning in both supervised and reinforcement learning settings \citep{maml}.
In the reinforcement learning (RL) setting, MAML learns a shared policy initialization across tasks.
This policy initialization is then adapted to new tasks at test time by policy gradient updates.

Transferring a policy between tasks faces two challenges. First, transferring a policy proves difficult on task distributions in which the optimal policies vary dramatically across tasks.
We must assume the existence of a policy initialization from which many task-specific policies may be found via a few local updates. This assumption may fail if policies vary widely across tasks.\footnote{While \citet{finn2017meta} show that a single step or multiple steps of gradient-based adaptation is powerful and universal in theory, this analysis may require very complex neural networks, which in turn need many samples to generalize.}

Second, transferring policies also affects the time and sample efficiency of the adaptation phase.
Even if an algorithm exhaustively trains against all tasks in the family, the adaptation phase still requires samples.

We aim to address these limitations by revisiting the model-based approach to multi-task reinforcement learning. First, we learn and adapt with a shared dynamical model, rather than a policy. Second, we propose a ``warm-up" phase of adaptation in which we train a policy on our learned dynamical model.
We assume that the tasks encountered have the same (or similar) associated dynamical models, but we allow the tasks, and the policies required to solve them, to vary arbitrarily.
Adaptation with our algorithm requires a modest amount of time (because we train a separate policy for each task), but the adaptation requires few, if any, samples.
Consider the extreme case in which we acquire a perfect dynamical model of the environment during training: adapting a policy to a new task may be computationally challenging, but will require no new samples.

Conventional wisdom may suggest the learned dynamical model will fail to transfer; i.e., a dynamical model learned for one task will not work for other tasks.
Perhaps surprisingly, we find that the dynamical model can transfer well; especially well if we also collect a small amount of new samples on a task during adaptation.
We find our model-based approach provides significant gains in sample complexity over prior state-of-the-art: policies produced by our algorithm obtain better performance on test tasks, despite using less than 1\% as much environment interaction at train time (e.g., 0.4 vs. 80 million samples).

In summary, we contribute:

\begin{enumerate}
    \item A model-based multi-task RL algorithm, Sequential Multi-task Learning (Algorithm \ref{alg:ours}).
    \item Numerical experiments comparing our approach to MAML on several continuous control MuJoCo reinforcement learning benchmarks of varying difficulty (Figure \ref{fig:comparisonresults}).
    \item Numerical experiments suggesting our algorithm can handle (a) out-of-distribution tasks, (b) a shift in dynamical model (both Figure \ref{fig:transferresults}) and (c) active task selection (Figure \ref{fig:activeresults}).
\end{enumerate}

\section{Preliminaries}

In multi-task learning we train on a variety of tasks to acquire some shared parameters that help us learn new, ``similar," tasks with few samples.
In this section we clarify the multi-task RL formalism addressed by our work, review the Model Agnostic Meta Learning (MAML) algorithm in this setting, and recall SLBO, a recent state-of-the-art model-based reinforcement learning algorithm.

\subsection{Reinforcement Learning}
Consider a Markov Decision Process (MDP) with state space $\CS$ and action space $\CA$. 
The transition dynamics $\dm(\cdot|s, a)$ specifies the conditional distribution of the next state given the current state $s$ and action $a$.
A reward function $r: \CS \times \CA \to \R$ defines the step-wise reward.
Additionally, we fix a discount $\gamma \in [0,1)$ and an initial state distribution $p_0$.

A policy $\pi(\cdot|s)$ specifies a conditional distribution over actions given a state $s$.
We define the value function $V^{\pi, M}: \CS \to \R$ at state $s$ for a policy $\pi$ and dynamical model $\dm$:
\begin{equation}
    V^{\pi,\dm}(s) = \underset{\substack{a_t \sim \pi(\cdot | s_t),\\ s_{t+1} \sim \dm(\cdot \mid s_t, a_t)}}{\BE} \left[
    \sum_{t = 0}^{\infty}\gamma^t r(s_t, a_t) \mid s_0 = s
    \right].
\end{equation}
In practice, we truncate the infinite sum to a finite horizon $H$.
The \textit{expected return} $\eta^{\dm}(\pi) := \underset{s_0 \sim p_0}{\BE}[V^{\pi,\dm}(s_0)]$ gives a one-number summary of a given policy's performance on an environment with dynamics $\dm$.
We seek a policy $\pi$ which maximizes $\eta^{\dm}(\pi)$.

\subsection{Multi-task Reinforcement Learning}
Let $\Psi \subseteq \R^k$ parameterize a family of \textit{tasks} and $\Theta \subseteq \R^p$ parameterize a family of policies.
The family of tasks, indexed by $\psi$, is a family of Markov decision process $\{(\CS, \CA, \dm_\psi, r_{\psi}, p_0, \gamma)\}_{\psi \in \Psi}$.
We are primarily interested in tasks that share (or have similar) underlying dynamics and only differ in their reward functions --- namely, the transition dynamics $\dm_\psi = \dm^\star$ for some fixed $\dm^\star$ (or $\dm_\psi$'s are similar). The value function of a policy $\pi$ on a task with reward $r_\psi$ and dynamics $\dm$ is denoted $V_\psi^{\pi,\dm}$, and the expected return $\eta_\psi^{\dm}(\pi) = \BE[V_\psi^{\pi,\dm}(s_0)]$ is defined now for each task and dynamical model.
To simplify notation, we will use the shorthand $\eta_\psi^{\dm}(\theta) := \eta_\psi^{\dm}(\pi_\theta)$, and omit the superscript $\dm$ in the common case where $\dm$ is the shared true dynamics $\dm^\star$.

To reuse knowledge across tasks, algorithms produce a shared \textit{structure}, such as a policy initialization (in MAML) or a dynamical model (in our algorithm).
Let $\Phi \subseteq \R^d$ denote the set of all such structures.
The training procedure produces one such shared structure $\phi \in \Phi$, which is subsequently used by an adaptation algorithm $A : \Phi \times \Psi \to \Theta$ at test time to produce policy parameters $\theta = A(\phi, \psi)$ for a given task $\psi$.
We consider a class of adaptation algorithms constrained to a common sample budget, but we suppress this budget in the notation for simplicity.

We aim to find a shared structure $\phi$ which enables good post-adaptation performance, in expectation, over a given task distribution $p(\psi)$:
\begin{equation}
    \maximize_{\phi}\; \underset{\psi}{\BE}\left[\,\eta_{\psi}(A(\phi, \psi))\,\right].
    \label{eq:metaobjective}
\end{equation}
If $A$ is non-deterministic, the expectation above is also taken over the randomness in $A$.

\subsection{Model Agnostic Meta Learning}

The celebrated Model Agnostic Meta Learning (MAML) algorithm, as applied to multi-task reinforcement learning, is a model-free policy-search initialization method \citep{maml}.
In MAML, the shared structure learned at train time is a set of policy parameters, i.e., $\Phi = \Theta$.
The adaptation algorithm updates the parameters of the policy for a new task so that they will perform well on the new task after a single step of gradient ascent (multiple steps may be used in practice):\begin{equation}
    A_{\textsc{maml}}(\theta, \psi) = \theta + \alpha \nabla \eta_{\psi}(\theta).
\end{equation}
where $\nabla \eta_{\phi}(\theta)$ is estimated via policy gradient methods \citep{williams1992simple, sutton2000policy} and $\alpha \in \R_+$ is a step size. The meta-objective, Eq. \ref{eq:metaobjective}, is approximately optimized over $\phi$ by sampling tasks.

\subsection{Stochastic Lower Bound Optimization}

The recently proposed Stochastic Lower Bound Optimization (SLBO) method is a model-based reinforcement learning algorithm \citep{luo2018algorithmic}.
SLBO achieves state-of-the-art sample efficiency by interleaving dynamical model fitting, policy training, and sample collection. 

For a learned dynamical model $\hat{\dm}_{\phi}$,
define the $k$-step prediction $\hat{s}_{t+k}$ on state $s_t$ and action sequence $a_{t:t+k}$ by $\hat{s}_t = s_t$ and $\hat{s}_{t+k+1}$ = $\hat{\dm}_\phi(\hat{s}_{t+k}, a_{t+k})$ for $k \geq 0$. Define the $k$-step prediction loss:
\begin{equation}
    \CL^{(k)}(s_{t:t+k}, a_{t:t+k}; \phi) := \frac{1}{k}\sum_{i = 1}^{k} \norm{\hat{s}_{t+i} - s_{t+i}}_{2}
    \label{eq:dynamicsloss}
\end{equation}

Consider a family of policies $\{\pi_{\theta}\}$ and dynamics models $\{\hat{\dm}_{\phi}\}$ (for our purposes, fully connected neural networks).
SLBO approximately solves: 
\begin{equation}
    \underset{\theta, \phi}{\text{maximize}}\;\left\{ \underbrace{\eta^{\hat\dm_{\phi}}(\pi_\theta)}_{\text{reward-to-go}} - \lambda \underbrace{\underset{s_{t:t+h}, a_{t:t+h} \sim \pi_{\tilde{\theta}}, \dm^\star}{\BE} \left[\CL^{(k)}(s_{t:t+k}, a_{t:t+k}; \phi) \right]}_{\text{transition dynamics fit}}\right\}
    \label{eq:slboobjective}
\end{equation}
by alternatively maximizing the reward-to-go on the virtual MDP induced by the learned dynamics $\hat\dm_\phi$, and minimizing the transition dynamics fit on data from a policy rollout.
Here $\lambda$ is a hyperparameter which controls how to weight the terms when optimizing and $\tilde{\theta}$ is the policy at the previous iteration. 
In practice, the transition dynamics are fit with Adam \citep{adam} and reward-to-go is optimized with TRPO \citep{schulman2015trust}.

\section{Related Work}
\label{sec:relatedwork}

A number of previous works have considered multi-task reinforcement learning \citep{oh2017zero, ammar2014online,wilson2007multi,lazaric2010bayesian}, and \citet{transferlearningsurvey} survey the area. 
One line of work studies the problem of producing a single policy which can effectively perform a variety of tasks.
A common strategy here, often referred to as \textit{distillation}, is to transfer knowledge from one policy to another.
The basic approach trains a single policy to imitate many task-specific expert policies \citep{actormimic,policydistillation}.
One can additionally regularize the task-specific policies so that they remain somewhat close to the distilled policy \citep{distral}.

Multi-task learning is closely related to \textit{meta-learning}, where a learning algorithm attempts to ``learn how to learn'', somehow leveraging prior experience to learn future tasks more quickly \citep{schmidhuber1987, thrun2012learning}.
Beyond reinforcement learning, meta-learning has been successfully applied to the problem of few-shot classification, where the goal is to learn (at test time) to accurately classify instances of new classes not encountered during training, given just a few examples from these classes \citep{santoro2016meta, vinyals2016matching, ravilarochelle}.
The MAML algorithm \citep{maml} approaches meta-learning by learning an initialization that can be quickly adapted to new tasks via gradient-based optimization.
Another approach is training a recurrent neural network that is a function of the training history, implicitly encoding a learning algorithm in its weights \citep{hochreiter2001learning}.
This strategy has been applied in reinforcement learning by \citet{rl2} and \citet{learn2rl}.
\citet{learn2adapt} develop both MAML-like and RNN-based meta-learning approaches to learning and adapting dynamical models online in changing environments.

Model-based approaches have long been recognized as a promising avenue for reducing the sample complexity of RL algorithms \citep{suttonbarto,pilco}.
We use an existing model-based RL algorithm (SLBO), noting that in principle our approach can be used with any model-based RL algorithm, and that SLBO is related to other recently proposed algorithms.
The special case where $n_{\text{inner}} = 1$ (i.e. the model is re-trained once each time new data is observed) has been described several times under different names: it is referred to as MB-TRPO in \citep{luo2018algorithmic}, as ``Vanilla Model-Based Deep Reinforcement Learning'' in \citep{metrpo}, and as SimPLe in \citep{mbatari}.
However, it is usually challenging to obtain a perfect dynamics model \citep{abbeel2006using}, and a policy trained on a single estimated model is prone to overfit to particular inaccuracies in that model.
\citet{metrpo} propose to use an ensemble of models to mitigate this issue.
Using multiple inner iterations in SLBO is similar to using an ensemble; due to stochasticity in the optimizer, the intermediate models obtained after varied numbers of optimization steps are likely to agree on the training data, but may differ outside of the training distribution.

Training a policy in a virtual environment is not the only way to use a learned dynamical model \citep{suttonbarto}.
For example, one can produce policies based on \textit{model predictive control} (MPC), where at each time step the model is used to perform planning over a short horizon to select the next action \citep{chua2018deep}.
\citet{nagabandi2018neural} use MPC to initalize policies and then fine-tune them using model-free RL.
Alternatively, one can use the model to generate ``imagined'' trajectories and use these as additional inputs to a policy \citep{i2a}.

\section{Our Approach}\label{sec:approach}

Recall that two areas of interest addressed by our approach are (a) sample complexity in the training time and adaptation time and (b) zero-shot adaptation to similar tasks. To achieve this we leverage two insights:
\begin{enumerate}
    \item We transfer the parameters of the learned \textit{dynamical model}, rather than the policy.
    \item We \textit{warm-up} a task's policy by training on learned ``virtual" dynamics prior to interaction.
\end{enumerate}

Our algorithm proceeds in three iterated stages: (1) task sampling, (2) policy warm-up training, and (3) one or more interleaved data collection and policy training phases.
In essence, we perform model-based reinforcement learning on a sequence of tasks sampled from the task distribution $p(\psi)$. 
We present details on each step below and pseudo-code in Algorithm \ref{alg:ours}. 

\subsection{Sequential Multi-task Training Overview}

\textbf{(1) Task Sampling.} We sample tasks from the distribution $p(\psi)$. 
We could augment this form of sampling by choosing tasks actively, or in some adversarial manner. Later we suggest some heuristics for choosing tasks actively (Section \ref{sec:activeapproach}), and we leave adversarial multi-task setting to later work.

\textbf{(2) Policy Warm-Up.} We initialize a random policy and warm it up on the new task by training on learned dynamical models; for the first task, we skip the warm-up.
We call this \texttt{VirtualTraining} because it requires no real samples, and trains the policy well against only learned dynamical models of the environment.
We intend, of course, that this warm-up procedure on the learned dynamical model will produce a reasonable policy without interacting with the new environment (validated empirically in Section \ref{sec:experiments}).

\textbf{(3) Data Collection.} Here we alternately collect new data, fit a dynamics model and then perform \texttt{VirtualTraining} for $n_{\text{slbo}}$ iterations. This stage is properly viewed as running SLBO, a model-based RL algorithm, on the new task with a warmed-up policy and previously selected data.
Only here (line \ref{alg:datacollection}) do we collect samples from the real environment.

\begin{algorithm}[H]
\SetAlgoLined
\SetKwFunction{VirtualTraining}{VirtualTraining}
\SetKwFunction{SLBO}{SLBO}
\KwResult{Model parameters $\phi$, dataset $\CD$}
\DontPrintSemicolon
    Initialize model parameters $\phi$ and dataset $\CD \leftarrow \emptyset$\;
    \For{$n_{\text{tasks}}$ iterations}{
        Sample task $\psi \sim p(\psi)$\;\label{alg:sampling}
        Initialize policy parameters $\theta$\;
        \text{If} $\CD \neq \emptyset$, \VirtualTraining{$\theta$, $\phi$, $\CD$, $n_{\text{warmup}}$}\;\label{alg:warmup}
        {(\SLBO):}\label{alg:slbo:start}
        \For{$n_{\text{slbo}}$ iterations}{
            $\CD \leftarrow \CD \cup \{$ collect $n_{\text{collect}}$ samples from real environment $\dm$ using $\pi_{\theta}$ with noise$ \}$\;\label{alg:datacollection}
            \VirtualTraining{$\theta$, $\phi$, $\CD$, $n_{\text{inner}}$}
        }\label{alg:slbo:end}
    }
\SetKwProg{subroutine}{subroutine}{ begin}{end}
\subroutine{\VirtualTraining{$\theta$: policy, $\phi$: model,  $\CD$: data, $n_{\text{inner}}$}\label{alg:vt:start}}{
    \For{$n_{\text{inner}}$ iterations}{
        Optimize Eq. \ref{eq:dynamicsloss} over $\phi$ with data sampled from $\CD$ by $n_{\text{model}}$ steps of Adam\;
        \For{$n_{\text{policy}}$ iterations}{
            $\CD' \leftarrow \{$ collect $n_{\text{trpo}}$ samples using learned $\hat{\dm}_{\phi}$ as dynamics $\}$\;
            Optimize $\pi_{\theta}$ by running TRPO on $\CD'$\;
        }
    }\label{alg:vt:end}
}
\caption{Sequential Multi-task Training}
\label{alg:ours}
\end{algorithm}

\subsection{Key Design Choices}
\label{sec:designchoices}

\textbf{Inner Iterations of Virtual Training.}
In steps (2) and (3), we train the policy against several learned dynamical models.
By intentional over-parameterization of the neural network representing the dynamical model, we cause the policy to see different dynamical models (all, however consistent with the data) at each iteration of the virtual training process.
Intuitively, this process prevents the policy from over-fitting to a particular dynamical model.
Empirically, \citep{luo2018algorithmic} found this improved performance in the single-task traditional RL setting. 

\textbf{Warm-up.} We warm-up a policy on the trained model. Firstly, this allows us to adapt the policy without any new samples. If we can collect new samples from the environment, this process ensures that we obtain informative data even on the first roll-out. Ultimately when we test our algorithm, our adaptation will be exactly this warm-up phase (followed, perhaps, by a few iterations of data collection and training phases).

\textbf{Sequential vs. Joint Training.} We have chosen to perform training \textit{sequentially}, meaning that we select a task and then train and collect data on it in isolation, before moving to the next task.
In contrast, we could instead sample a batch of tasks, then iteratively collect data and train policies for each task \textit{jointly}.
The primary advantage of sequential training is that for later tasks, we start with a reasonable policy, and so data collected is very informative.
Whereas in joint training, we waste samples early on by collecting data on all tasks when each policy is poor.

\subsection{Adaptation / Test Phase}

Given a new task, we use the learned model to train a policy from scratch.
More precisely, we randomly initialize policy parameters $\tilde\theta_0$, warm the policy up using $\hat{\dm}_\phi$, and then continue to adapt with further SLBO training:
\begin{equation}
   A_{\textsc{ours}}(\phi, \psi) = \textsc{slbo}(\textsc{vt}(\tilde\theta_0, \phi, \CD, n_{\text{warmup}}), \phi, \CD, n_{\text{slbo}}).
   \label{ref:oursadaptation}
\end{equation}
where $\textsc{vt}$ denotes the \texttt{VirtualTraining} routine (lines \ref{alg:vt:start}-\ref{alg:vt:end} in Algorithm \ref{alg:ours}) which returns updated policy parameters, and $\textsc{slbo}$ denotes some number of steps of the \texttt{SLBO} algorithm (lines \ref{alg:slbo:start}-\ref{alg:slbo:end} in Algorithm \ref{alg:ours}).

In Section \ref{sec:shift} we experimentally analyze our algorithm's performance when adapting to (a) different task distributions at test time and (b) different dynamical models at test time.

\subsection{Active Task Selection}
\label{sec:activeapproach}

As mentioned earlier, our tasks need not be sampled uniformly.
In particular, Line \ref{alg:sampling} of Algorithm \ref{alg:ours} can be extended to select tasks actively.
This enables us to select a particularly difficult or diverse sequence of tasks, which may be desirable if it speeds training.

\textbf{Skipping Tasks.} Suppose we have a function $\mu: \Psi \to \R$ that faithfully rates the difficulty of a task:
$(\psi_1 \text{ harder than } \psi_2) \Longleftrightarrow \mu(\psi_1) > \mu(\psi_2)$ where the LHS is domain specific. If we have seen at least $L$ tasks, a straightforward active augmentation to Algorithm \ref{alg:ours} at iteration $j > L$ is to skip task $\psi_j$ if $\mu(\psi_{j}) < \operatorname{quantile}_{q}(\{\mu(\psi_{i})\}_{i=1}^{L})$.
Intuitively, we want to focus on tasks which are ``hard."

\textbf{Rating Tasks.} One sensible rating function is the difference between a virtual model's predicted policy reward and the real reward:
\[
    \mu(\psi_i) = \eta^{\hat{M}_{\phi}}_{\psi_i}(\theta_i) - \eta^{M^*}_{\psi_i}(\theta_i),
\]
where $\theta_i$ are parameters of the policy after warm-up.
Computing such a rating requires samples, but we can estimate it using several pairs of virtual models. We experimentally explore active task sampling in Section \ref{sec:active}.

\section{Experiments}\label{sec:experiments}

\begin{figure}
    \centering
    \begin{subfigure}[b]{0.23\textwidth}
        \centering
        \includegraphics[width=\textwidth]{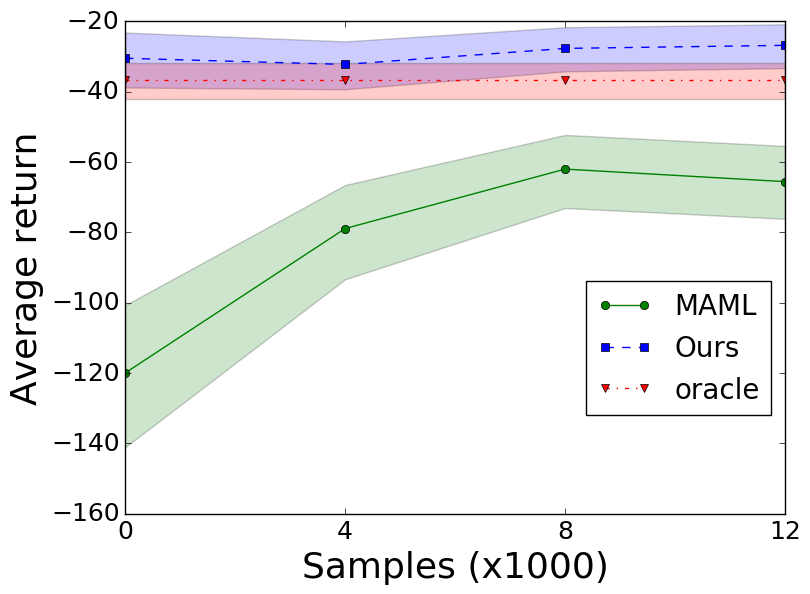}
        \caption{Cheetah $[0, 2]$}
    \end{subfigure}
    \hfill
    \begin{subfigure}[b]{0.23\textwidth}
        \centering
        \includegraphics[width=\textwidth]{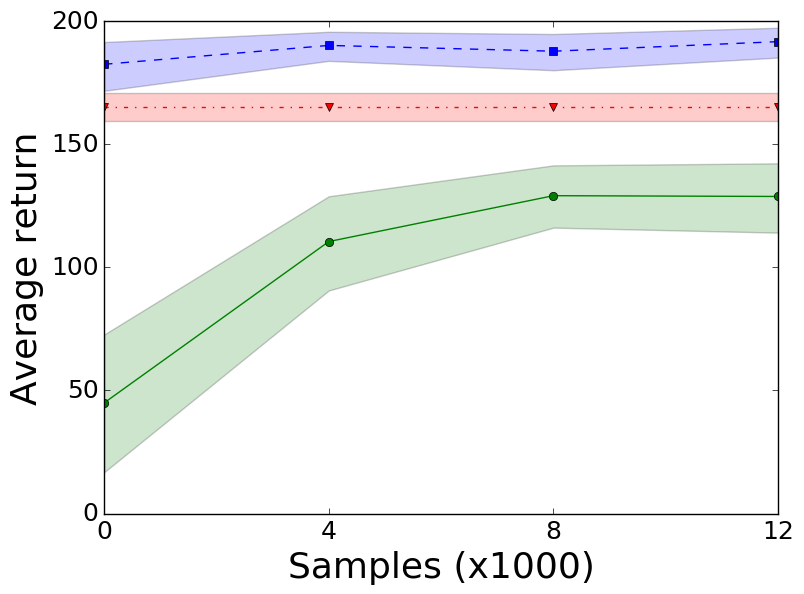}
        \caption{Ant $[0, 3]$}
    \end{subfigure}
    \hfill
    \begin{subfigure}[b]{0.23\textwidth}
        \centering
        \includegraphics[width=\textwidth]{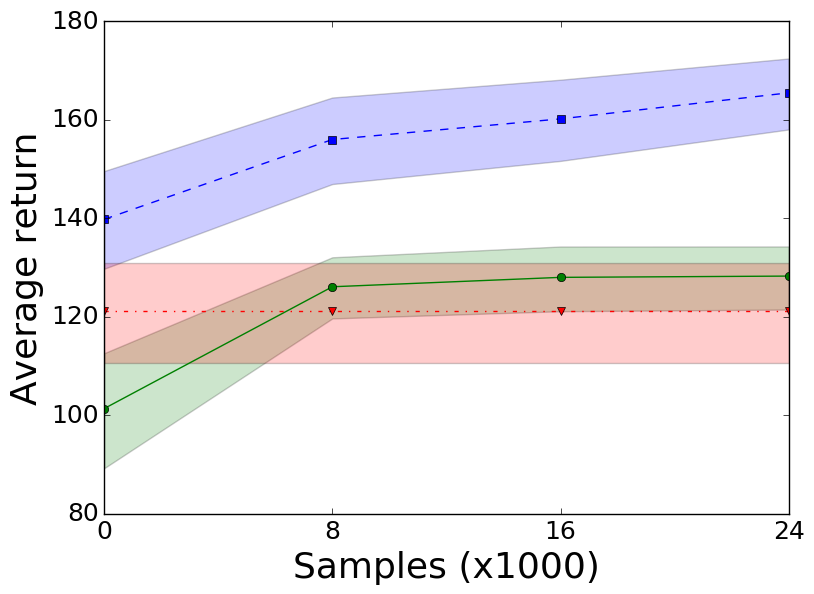}
        \caption{Humanoid $[0, 1.5]$}
    \end{subfigure}
    \hfill
    \begin{subfigure}[b]{0.23\textwidth}
        \centering
        \includegraphics[width=\textwidth]{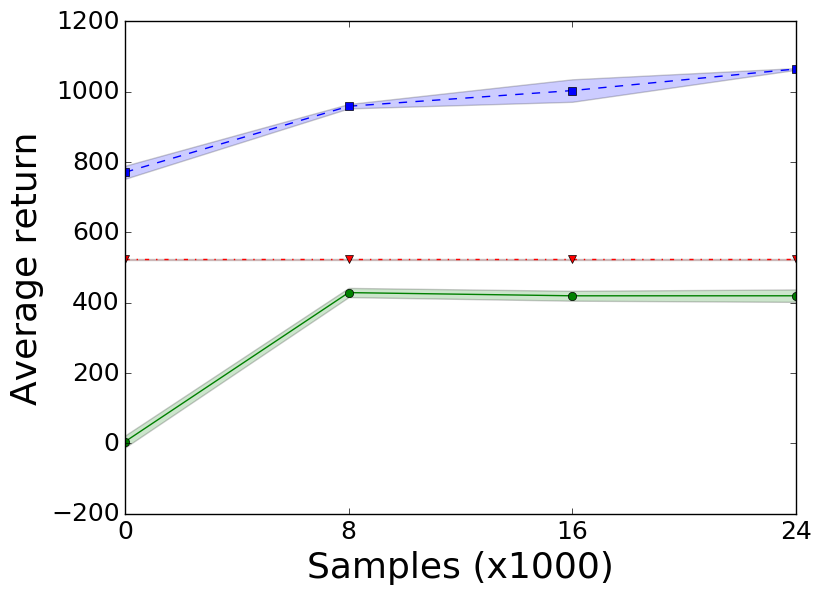}
        \caption{Ant forward/back}
        \label{fig:antforwardbackward}

    \end{subfigure}
    \hfill
    \vfill
    \hfill
    \begin{subfigure}[b]{0.23\textwidth}
        \centering
        \includegraphics[width=\textwidth]{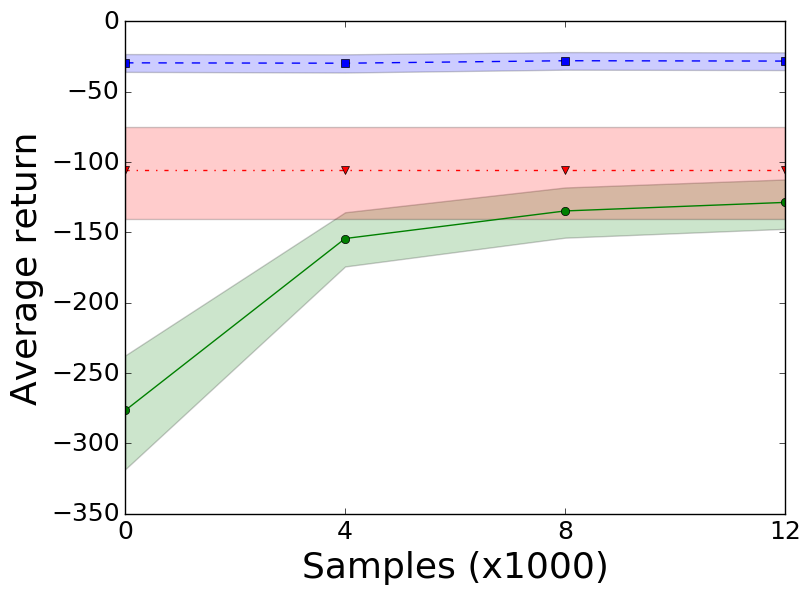}
        \caption{Cheetah $[-2, 2]$}
        \label{fig:cheetahfull}
    \end{subfigure}
    \hfill
    \begin{subfigure}[b]{0.23\textwidth}
        \centering
        \includegraphics[width=\textwidth]{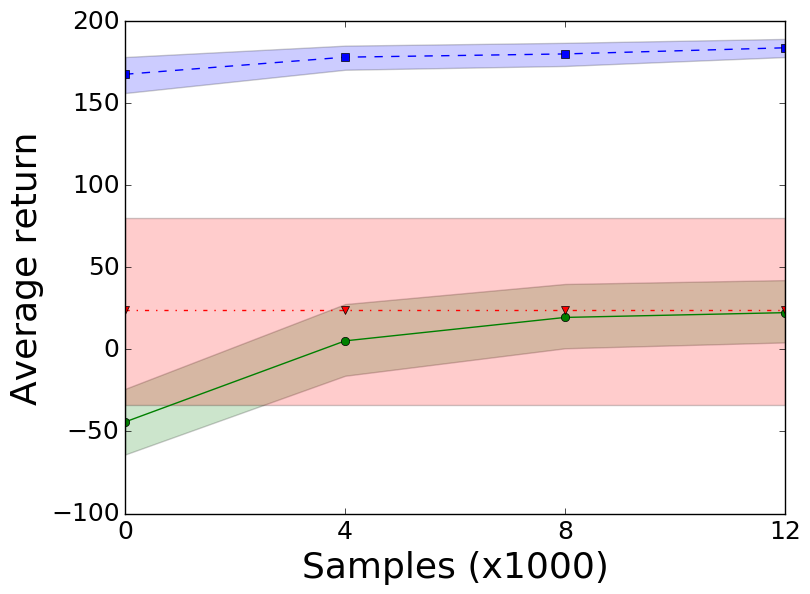}
        \caption{Ant $[-3, 3]$}
        \label{fig:antfull}
    \end{subfigure}
    \hfill
    \begin{subfigure}[b]{0.23\textwidth}
        \centering
        \includegraphics[width=\textwidth]{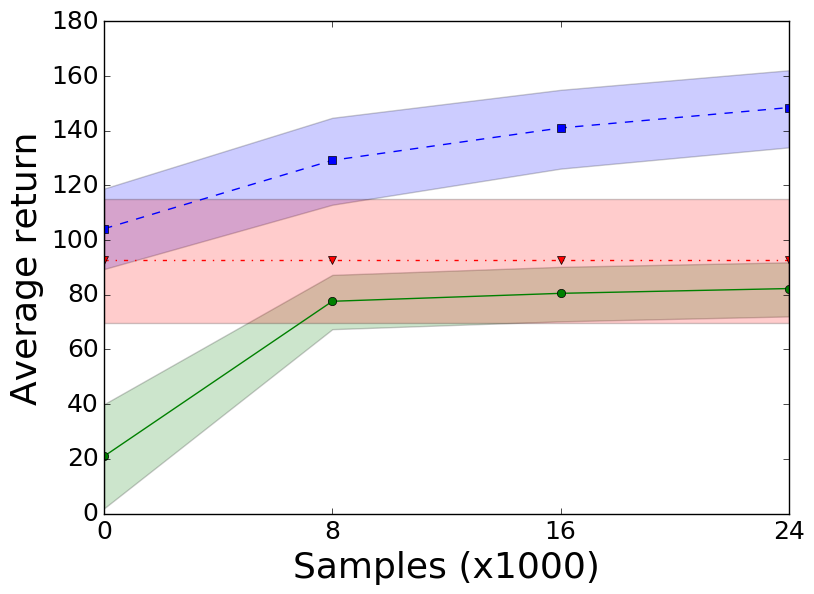}
        \caption{Humanoid $[-1.5, 1.5]$}
    \end{subfigure}
    \hfill
    \begin{subfigure}[b]{0.23\textwidth}
        \centering
        \includegraphics[width=\textwidth]{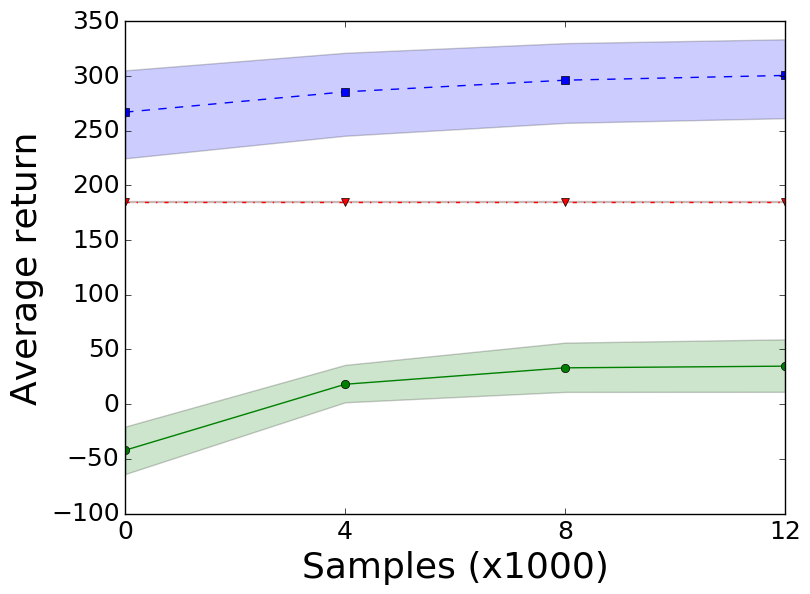}
        \caption{Ant $[-3, 3]^2$}
    \end{subfigure}
    \hfill
    \caption{Average return (with 95\% confidence interval) across tasks of adapted policies from our algorithm, MAML, and an ``oracle" policy trained with access to task parameters as input. 
        We substantially outperform MAML, often with no interaction; a result of transferring a dynamical model and warm-up.
    See Section \ref{sec:maincomparison}.}                 \label{fig:comparisonresults}
    \end{figure}

We evaluate our approach on a variety of continuous control tasks based on environments from the rllab benchmark \citep{rllab}, which uses the MuJoCo physics simulator \citep{mujoco}.
We first describe the tasks considered, explain the setup of the experiments, present results comparing our algorithm and MAML.
Next, we describe and give results for modified settings where the test tasks are not drawn from the same distribution as the training tasks.
Finally, we include preliminary results for methods in which we sample tasks adaptively, rather than independently, at train time.

We implemented our sequential multi-task training algorithm by extending the code that \citet{luo2018algorithmic} provide.\footnote{\url{https://github.com/roosephu/slbo}}
We use the MAML implementation that \citet{maml} provide. \footnote{\url{https://github.com/cbfinn/maml_rl}}
In all cases, we use a horizon of $H = 200$ time steps. Each task involves one of three standard MuJoCo models: \textit{half-cheetah}, \textit{ant}, or \textit{simple humanoid}.

\label{sec:taskdescriptions}

\textbf{Goal velocity tasks.}
In \textit{goal velocity} tasks, a velocity $\psi$ is given, and the agent's goal is to match its own velocity to $\psi$.
In our experiments, $\psi$ is sampled uniformly at random from an interval $\Psi$.
We experimented with the following variants: half-cheetah with $\Psi = [0,2]$, half-cheetah with $\Psi = [-2,2]$, ant with $\Psi = [0,3]$, ant with $\Psi = [-3,3]$, humanoid with $\Psi = [0,1.5]$, humanoid with $\psi = [-1.5,1.5]$, and ant with $x$ and $y$ velocities $\Psi = [-3, 3]^2$.\footnote{$[0,b]$ and forward/backward tasks first appeared in \citet{maml}. We remove contact cost, a negligible part of total reward, as we do not include this in  our learned dynamical model.}

\textbf{Forward/backward tasks.}
In \textit{forward/backward} tasks, a direction $\psi \in \{\pm 1\}$ (where $\psi = 1$ indicates forward and $\psi = -1$ indicates backward) is given, and the agent's goal is to maximize its speed in the given direction.
In our experiments, $\psi$ is sampled uniformly at random from $\Psi = \{\pm 1\}$.

\subsection{Training \& Evaluation}\label{sec:traineval}
We compare the performance of our method to MAML in adapting to new tasks sampled from the task distribution.
The output of MAML's meta-training process is a set of initial policy parameters $\theta_0$, while the output of our method is a set of initial dynamical model parameters $\phi_0$ and a dataset $\CD_0$.

\textbf{Training.} The training process for our algorithm is outlined in Algorithm \ref{alg:ours}. The training process for MAML is outlined in \citet{maml}, Algorithm 1.
For both algorithms, and all environments, each policy is implemented as a fully connected feedforward network with two hidden layers of 100 units each, using the ReLU activation $\sigma(x) = \max\{0,x\}$.
The sizes of the input and output layers are matched with the dimensions of the state and action spaces, respectively.
We list all the hyperparameters used for both algorithms in the Appendix.

We note here that the MAML training process (which we did not modify from \cite{maml}) requires 80 million samples. We ran our algorithm for $n_{\text{tasks}} = 100$ iterations, requiring 0.4 million samples, less than 1\% required by MAML\footnote{Ours: 100 tasks by 4000 samples/task. MAML: 500 meta-steps by 40 tasks/meta-step by 4000 samples/tasks.}.

\textbf{Evaluation.} To evaluate our algorithm against MAML we sample $40$ tasks from $p(\psi)$ and train a separate policy for each.
For our algorithm we warm-up a policy on the virtual environment $\phi_0$ using data sampled from $\CD_0$, and then train on $n_{\text{itr}} = 3$ stages of SLBO.
For MAML, we initialize to the policy $\theta_0$ and then train $n_{\text{itr}} = 3$ policy gradient steps.
In both cases, the amount of data collected at each iteration is the same, and is plotted on the horizontal axis of all plots.

\subsection{Comparison to MAML}
\label{sec:maincomparison}

For the eight task/environment pairs described in Section \ref{sec:taskdescriptions}, we compare our algorithm against (a) MAML and (b) an \textit{oracle} policy which is trained jointly across the task distribution but receives the task parameter $\psi$ as an additional input (as in \citep{maml}).
We produce the curves in Figure \ref{fig:comparisonresults} by (1) sampling 40 independently drawn tasks, (2) training a policy by collecting roll-outs (described in Section \ref{sec:traineval}), and (3) estimating the policy's average return by sampling new roll-outs.
The plots show the return averaged across tasks, with 95\% confidence intervals computed by a bootstrap estimator.
Although different tasks have different reward ranges (i.e., the maximum achievable reward depends on the task parameter $\psi$), the expectation over tasks is comparable. The oracle policy acts roughly as an upper bound of the few-step performance of MAML, but we note that our approach can and often does exceed the performance of the oracle policy, because we train a separate, independent policy for each test task, whereas the oracle policy takes in the task parameters as inputs.
MAML also trains a policy for each test task, but these are all constrained to have the same initialization, effectively limiting their ability to cover the task distribution.

\textbf{Analysis:}  
Firstly, we out-perform MAML across all tasks; even having trained with fewer samples. As we mentioned earlier, if the policy must vary substantially on different tasks within the same task family, there is no guarantee that there exists an initialization from which near-optimal policies for all tasks can be reached within one or a few gradient steps. This is a fundamental limitation of MAML which is not shared by our approach, and allows us to substantially out-perform MAML.
Secondly, training policies in a virtual environment provided by the learned dynamics model enables \textit{zero-shot adaptation} to new tasks if the model is sufficiently accurate.
That is, the policy produced by the warm-up stage (before any samples are collected from the test environment) may already be high-performing.
Indeed, we observe our algorithm out-performing the oracle without any samples.

\subsection{Task Distribution and Dynamical Model Shift}\label{sec:shift}
\begin{figure}
     \centering
     \begin{subfigure}[b]{0.23\textwidth}
         \centering
         \includegraphics[width=\textwidth]{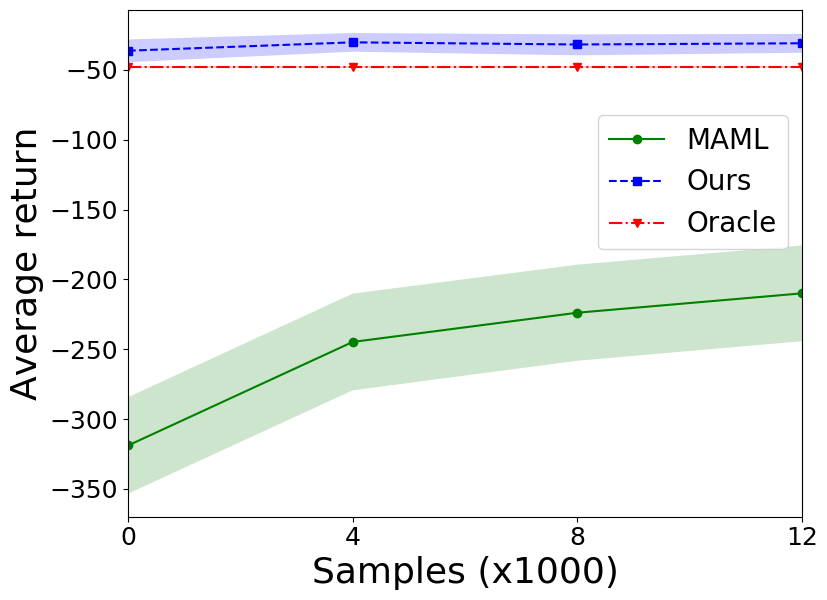}
         \caption{Cheetah Pos-to-Neg}
         \label{fig:cheetahposneg}
     \end{subfigure}
     \hfill
     \begin{subfigure}[b]{0.23\textwidth}
         \centering
         \includegraphics[width=\textwidth]{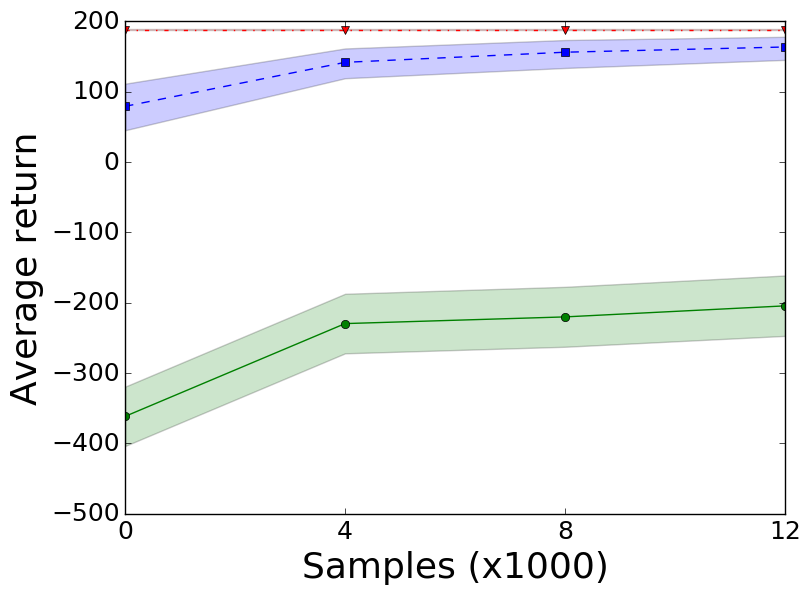}
         \caption{Ant Pos-to-Neg}
        \label{fig:antposneg}
     \end{subfigure}
     \hfill
     \begin{subfigure}[b]{0.23\textwidth}
         \centering
         \includegraphics[width=\textwidth]{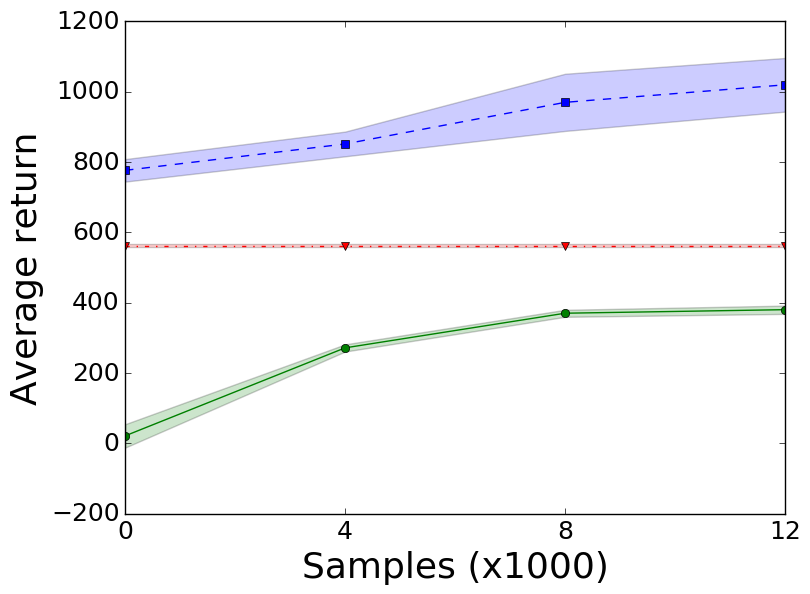}
         \caption{Ant low friction}
         \label{fig:antlowfric}
     \end{subfigure}
     \hfill
    \hfill
    \begin{subfigure}[b]{0.23\textwidth}
         \centering
         \includegraphics[width=\textwidth]{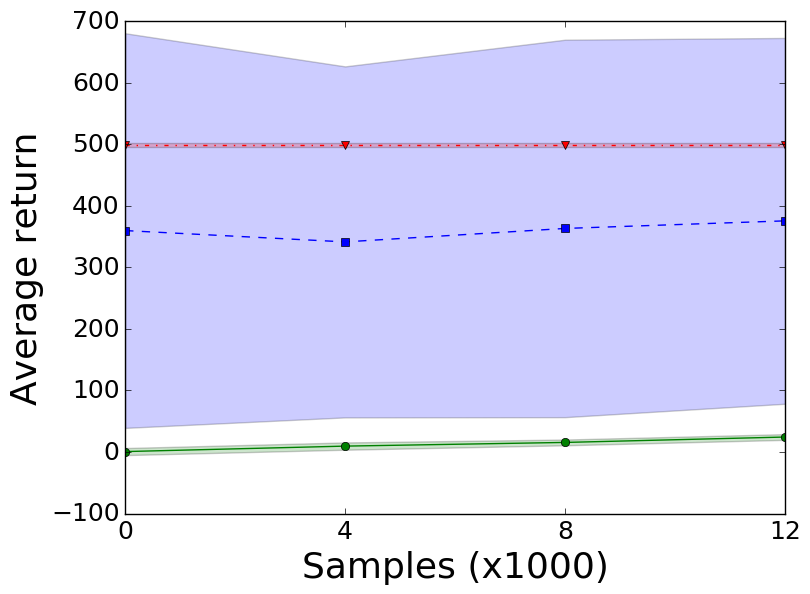}
         \caption{Ant crippled}
         \label{fig:antcrippled}
     \end{subfigure}
    \hfill
    \caption{Average return (with 95\% confidence interval) across tasks of adapted policies from our algorithm, MAML and the oracle baseline (Section \ref{sec:maincomparison}) for task distribution and dynamical model shift.
    Even on unseen tasks, warm-up performs well. When underlying dynamical model changes, we (naturally) require some samples to adapt (c) and (d). See Section \ref{sec:shift} for details.}
    \label{fig:transferresults}
    \end{figure}

We also examine how our approach performs when the distribution of tasks at test time differs from that at train time.
This scenario is referred to as \textit{domain adaptation} in the context of supervised learning \citep{bd2010}. 
We experimented with four such tasks, results in Figure \ref{fig:transferresults}.

\textbf{Changing Reward Distribution.} Our first pair of set-ups (one for half-cheetah, and one for ant) evaluate transfer to different reward distributions, without changing the dynamics.
In these \textit{positive to negative} tasks, we train on positive velocities ($[0,2]$ for half-cheetah, $[0,3]$ for ant) but we test on velocities from the negated interval (i.e., $[-2,0]$ and $[-3,0]$, respectively).

\textbf{Changing True Dynamical Model.} Our second two set-ups maintain the same reward distribution, but evaluate transfer to somewhat different underlying dynamical models.
In the \textit{ant low friction} scenario, the setup we train on the usual \textit{ant forward/backward} task family, but test in a MuJoCo environment with all friction coefficients halved. 
In the \textit{ant crippled} scenario, we train on the usual \textit{ant forward/backward} task family, but test in a MuJoCo environment with one ant leg disabled.

\textbf{Analysis:} We see that in the changing reward distributions (Figures \ref{fig:cheetahposneg}, \ref{fig:antposneg}), our algorithm is able to still achieve some gains over MAML as a result of the warm-up phase (compare with Figures \ref{fig:cheetahfull} and \ref{fig:antfull}). As expected, if the underlying dynamical model changes (i.e., the friction in Figure \ref{fig:antlowfric}) our warm-up is not as effective. Especially so in the case when we cripple the ant, compare the mean reward in Figure \ref{fig:antcrippled} (reward \textasciitilde$400$) with that in Figure \ref{fig:antforwardbackward} (reward \textasciitilde$1000$).

\subsection{Active Task Selection}
\label{sec:active}
\begin{figure*}
\centering
  \hfill
    \begin{subfigure}[b]{0.32\textwidth}
         \centering
         \includegraphics[width=\textwidth]{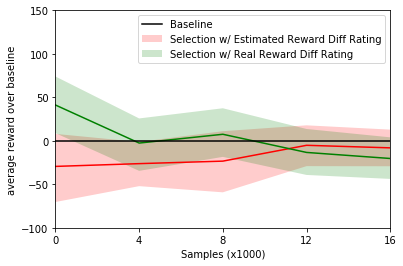}
         \caption{5 tasks sampled}
     \end{subfigure}
    \hfill
      \hfill
    \begin{subfigure}[b]{0.32\textwidth}
         \centering
         \includegraphics[width=\textwidth]{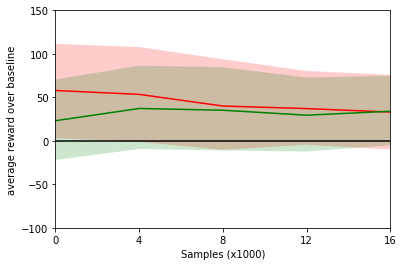}
         \caption{20 tasks sampled}
     \end{subfigure}
    \hfill
      \hfill
    \begin{subfigure}[b]{0.32\textwidth}
         \centering
         \includegraphics[width=\textwidth]{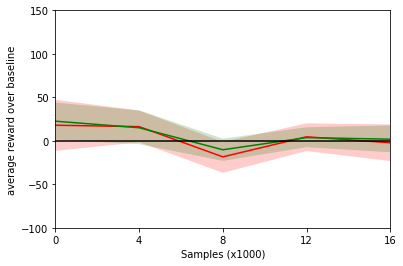}
         \caption{30 tasks sampled}
     \end{subfigure}
    \hfill
    \caption{Average difference in return (with 95\% confidence intervals) of our active algorithm vs. our non-active algorithm as a baseline. For \textit{ant velocity} $[-3, 3]^2$ environment. We show results for two task difficulty rating functions: estimated (no new samples) and true predicted virtual vs. real policy reward. Active sampling helps in regime of 20 samples (positive return difference with baseline). See Section \ref{sec:active} for details.}
    \label{fig:activeresults}
\end{figure*}

We validate two instances of the active task selection algorithm described in (Section \ref{sec:activeapproach}): one using the real reward difference rating (requiring extra samples) and one estimating the reward difference (requiring no samples).
Figure \ref{fig:activeresults} shows the advantage over the non-active Algorithm \ref{alg:ours} on the two-dimensional ant velocity task in three regimes: 5, 20 and 30 tasks.

\textbf{Analysis:} We see that active sampling can not help much with only 5 tasks, helps a modest amount with 20 tasks, and by 30 tasks, our baseline is already doing so well that the active sampling no longer helps.
These results suggest the promise of active sampling, but also suggest that the tasks considered in this work are too easy to see any significant gains; our non-active algorithm works well already. 
We leave it to future work to explore harder tasks and different rating functions.

\section{Conclusion}

\textbf{Limitations.}
Of course, our ability to outperform MAML is enabled by the underlying assumption of similar dynamical models across tasks. 
In many applications, e.g. robotics, this assumption is reasonable.
Still, we do not pretend to solve the general problem when the transition dynamics vary significantly among the tasks to be learned.
In that case, zero-shot adaptation becomes impossible, but a partial dynamical model may accelerate adaptation.
Second, virtual training, although it requires no new samples, is compute-intensive and the warm-up phase proves the longest part of our algorithm.
Though the trade-off with samples is favorable: according to our rough measurements, our less sample-intensive training process still requires less wall-time than MAML. 

\textbf{Future directions.} Three clear directions lie ahead. (1) Apply Sequential Multi-task Learning to more difficult settings. In fact, we perform well on most setups in Section \ref{sec:experiments} within 30 tasks. Also, harder settings would allow us to evaluate active task selection algorithms.
(2) Sample tasks \textit{actively} from the distribution. We sketched one possibility in Section \ref{sec:active}, but upon reaching a sample budget of 30 we observe vanishing gains from active sampling. The questions of rating and comparing tasks, especially with no new samples, remain open.
(3) Analyze the online setting and adversarial tasks.
Our algorithm handles tasks \textit{sequentially} so the tasks need not come from any distribution.
Analysing worst case adaptation performance would be interesting.

\textbf{Summary.}
Our model-based approach to multi-task reinforcement learning (a) delivers sample efficiency and (b) enables zero-shot adaptation. 
We transfer the shared structure of a learned dynamical model and adapt to a new task by training on this learned dynamical model.
Our empirical results support the effectiveness of model-based multi-task reinforcement learning.

\clearpage
\bibliographystyle{plainnat}
\bibliography{biblio}

\clearpage
\section{Appendix}
\subsection{Hyperparameters}
For the purpose of reproducibility, we list here all the hyperparameters used in our experiments.

We first describe the settings used for our algorithm.
We parameterized the dynamical model by a simple feedforward neural network with two hidden layers of width 500. We used Adam to optimize the reconstruction loss, with a learning rate of 0.001 and a training bath size of 128. We parameterized the policy by a simple feed-forward neural network with two hidden layers of width 32. We used TRPO to optimize the policy with standard parameters from \cite{luo2018algorithmic}. As mentioned earlier, most experiments (except notably the active task selection ones) trained against 100 tasks (i.e., $n_{\text{tasks}} = 100$). Additionally, $n_{\text{warmup}} = 40$, $n_{\text{model}} = 100$, $n_{\text{policy}} = 40$ and $n_{\text{slbo}} = 1$ (i.e., we only collected samples once for each task after warm-up).  For the adaptation phase, of course, $n_{\text{slbo}} = 4$. On half cheetah environment we collected 4000 samples from the real environment per iteration (with a horizon of 200, this corresponds to 20 trajectories on average. For ant and humanoid we collected 8000 samples from the real environment per iteration (with a horizon of 200, this corresponds to 40 trajectories on average).  We always collected the same number of samples when performing TRPO on the virtual dynamics (i.e., $n_{\text{trpo}} = 4000$ or $8000$ respectively). We collect the same number of samples at adaptation time.

We now describe the settings used when evaluating MAML.
They were mostly taken directly from \citet{maml} or their supplied implementation.
The adaptation steps are computed by standard policy gradients \citep{williams1992simple}, while the meta-updates are computed by TRPO \citep{schulman2015trust}.
We performed $500$ iterations of MAML training.
The learning rate (referred to as $\alpha$ in \citet{maml}) is set to $0.1$ during the meta-train phase; during the meta-test phase, $\alpha = 0.1$ for the first step and $\alpha = 0.05$ for subsequent steps.
The meta-learning rate (referred to as $\beta$ in \citet{maml}) is set to $0.01$.
The batch size (the number of rollouts used to compute the policy gradient updates) is $20$, except for the ant forward/backward and all humanoid tasks, where we use $40$. 
The meta-batch size (the number of tasks sampled at each iteration of MAML) is $40$ in all cases.

\end{document}